\documentclass[11pt,a4paper]{article}
\usepackage[hyperref]{eacl2021}
\usepackage{times}
\usepackage{latexsym}
\usepackage{booktabs}
\usepackage{amsmath}
\usepackage{amssymb}
\usepackage{amsfonts}
\usepackage{graphicx}
\usepackage{tabularx}
\usepackage{multicol}
\usepackage{multirow}
\usepackage{arydshln}
\usepackage{url}
\usepackage{adjustbox}
\usepackage{CJKutf8}

\usepackage{microtype}

\aclfinalcopy 

\newcommand{\en}{\textsc{en}\,}
\newcommand{\tr}{\textsc{tr}\,}
\newcommand{\de}{\textsc{de}\,}
\newcommand{\fr}{\textsc{fr}\,}
\newcommand{\zh}{\textsc{zh}\,}
\newcommand{\norm}[1]{\left\lVert#1\right\rVert}

\title{Is Supervised Syntactic Parsing Beneficial for Language Understanding Tasks? An Empirical Investigation}

\author{Goran Glava\v{s} \\
  University of Mannheim \\
  Data and Web Science Group \\
  \texttt{goran@informatik.uni-mannheim.de} \\\And
  Ivan Vuli\'{c} \\
  University of Cambridge \\
  Language Technology Lab \\
  \texttt{iv250@cam.ac.uk} \\}

\date{}

\begin{document}
\maketitle
\begin{abstract}

Traditional NLP has long held (supervised) syntactic parsing necessary for successful higher-level semantic language understanding (LU). The recent advent of end-to-end neural models, self-supervised via language modeling (LM), and their success on a wide range of LU tasks, however, questions this belief.
In this work, we empirically investigate the usefulness of supervised parsing for semantic LU in the context of LM-pretrained transformer networks.  
Relying on the established fine-tuning paradigm, we first couple a pretrained transformer with a biaffine parsing head, aiming to infuse explicit syntactic knowledge from Universal Dependencies treebanks into the transformer. We then fine-tune the model for LU tasks and measure the effect of the intermediate parsing training (IPT) on downstream LU task performance. Results from both monolingual English and zero-shot language transfer experiments (with intermediate target-language parsing) show that explicit formalized syntax, injected into transformers through IPT, has very limited and inconsistent effect on downstream LU performance.        
Our results, coupled with our analysis of transformers' representation spaces before and after intermediate parsing, make a significant step towards providing answers to an essential question: how (un)availing is supervised parsing for high-level semantic natural language understanding in the era of large neural models?

\end{abstract}

\section{Introduction}
\label{sec:intro}
Structural analysis of sentences, based on a variety of syntactic formalisms \cite[\textit{inter alia}]{charniak1996tree,taylor2003penn,de2006generating,hockenmaier2007ccgbank,nivre2016universal,nivre2020universal}, has been the beating heart of NLP pipelines for decades \cite{klein2003accurate,chen2014fast,dozat2016deep,kondratyuk201975}, establishing rather strong common belief that high-level semantic language understanding (LU) crucially depends on explicit syntax.   
The unprecedented success of neural language learning models based on transformer networks \cite{vaswani2017attention}, trained on unlabeled corpora via language modeling (LM) objectives \cite[\textit{inter alia}]{devlin2019bert,liu_roberta_2019,clark2020electra} on a wide variety of LU tasks \cite{wang-etal-2018-glue,hu2020xtreme}, however, questions this widely accepted assumption. 

The question of necessity of supervised parsing for LU and NLP in general has been raised before. More than a decade ago, \newcite{bod2007end} questioned the superiority of supervised parsing over unsupervised induction of syntactic structures in the context of statistical machine translation. Nonetheless, the NLP community has since still managed to find sufficient evidence for the usefulness of explicit syntax in higher-level LU tasks \cite[\textit{inter alia}]{levy2014dependency,cheng2015syntax,bastings2017graph,kasai2019syntax,zhang-etal-2019-syntax}. However, we believe that the massive improvements brought about by the LM-pretrained transformers -- unexposed to any explicit syntactic signal -- warrant a renewed scrutiny of the utility of supervised parsing for high-level language understanding.\footnote{\textbf{Disclaimer 1:} In this work, we make a clear distinction between Computational Linguistics (CL), i.e., the area of linguistics leveraging computational methods for analyses of human languages and NLP, the area of artificial intelligence tackling human language in order to perform intelligent tasks. This work scrutinizes the usefulness of supervised parsing and explicit syntax only for the latter. We find the usefulness of explicit syntax in CL to be self-evident.}\textsuperscript{,}\footnote{\textbf{Disclaimer 2:} The purpose of this work is definitely not to invalidate the admirable efforts on syntactic annotation and modeling, but rather to make an empirically driven step towards a deeper understanding of the relationship between LU and formalised syntactic knowledge, and the extent of its impact to modern semantic LU and applications.} 
The research question we address in this work can be summarized as follows: 

\vspace{1mm}
\noindent \textbf{(RQ)} \textit{Is explicit structural language information, provided in the form of a widely adopted syntactic formalism (Universal Dependencies, UD) \cite{nivre2016universal} and injected in a supervised manner into LM-pretrained transformers beneficial for transformers' downstream LU performance?}
\vspace{1mm}

\noindent While existing body of work \cite{lin2019open,tenney2019you,liu2019linguistic,kulmizev2020neural,chi2020finding} probes transformers for structural phenomena, our work is more pragmatically motivated. We directly evaluate the effect of infusing structural language information from UD treebanks, via intermediate dependency parsing (DP) training, on transformers' performance in downstream LU. To this end, we couple a pretrained transformer with a biaffine parser similar to \newcite{dozat2016deep}, and train the model (i.e., fine-tune the transformer) for DP. Our parser on top of RoBERTa \cite{liu_roberta_2019} and XLM-R \cite{Conneau:2020acl} produces DP results which are comparable to state of the art. 
We then fine-tune the syntactically-informed transformers for three downstream LU tasks: natural language inference (NLI) \cite{williams2018broad,conneau2018xnli}, paraphrase identification \cite{zhang2019paws,yang2019paws}, and causal commonsense reasoning   \cite{Sap:2019siqa,ponti2020xcopa}. We quantify the contribution of explicit syntax by comparing LU performance of the transformer exposed to intermediate parsing training (IPT) and its counterpart directly fine-tuned for the downstream task. We investigate the effects of IPT (1) \textit{monolingually}, by fine-tuning English transformers, BERT and RoBERTa, on an English UD treebank and for (2) downstream \textit{zero-shot language transfer}, by fine-tuning massively multilingual transformers (MMTs) -- mBERT and XLM-R \cite{Conneau:2020acl} -- on treebanks of downstream target languages, before the downstream fine-tuning on source language (English) data. 

While intermediate parsing training is obviously not the only way of bringing syntactic knowledge to downstream tasks \cite{kuncoro2019scalable,swayamdipta2019shallow,kuncoro2020syntactic}, it is arguably the most straightforward way of injecting syntactic signal in the context of the predominant pretraining-fine-tuning paradigm that has, nonetheless, not been investigated up to this point. Other methods of bringing syntactic signal to downstream tasks such as knowledge distillation \cite{kuncoro2020syntactic} and pre-training on shallow trees instead of sequences \cite{swayamdipta2019shallow} have failed to demonstrate significant gains on higher-level LU tasks.   

Our results also render supervised UD parsing largely inconsequential to LU. We observe limited and inconsistent gains only in zero-shot downstream language transfer: further analyses reveal that (1) intermediate LM training yields comparable gains and (2) IPT only marginally changes representation spaces of transformers exposed to sufficient amount of language data in LM-pretraining. We hope that these empirical findings will shed new light on the relationship between supervised parsing (and manually labeled treebanks) and LU with transformer networks, and guide further similar investigations in future work, in order to fully understand the impact of formal syntactic knowledge on LU performance with modern neural architectures.

\section{Related Work}

\noindent \textbf{Bringing Explicit Syntax to LMs.} Previous work has attempted to enrich language models with explicit syntactic knowledge in ways other than intermediate parsing training. \newcite{swayamdipta2019shallow} modify the pretraining objective of ELMo \cite{peters2018deep} to learn from shallowly parsed (i.e., chunked) corpora. They, however, report no notable improvements on downstream tasks. \newcite{kuncoro2019scalable} propose to distil the knowledge from a Recurrent NN Grammar (RNNG) teacher trained on a small syntactically annotated corpus (by modeling the joint probability of surface sequence and phrase structure tree) into an LSTM-based student pretrained on a much larger corpus. They show that distillation helps the student in structured prediction tasks, but their downstream evaluation does not involve LU tasks. Their subsequent work \cite{kuncoro2020syntactic} replaces the RNN student with BERT \cite{devlin2019bert}: syntactic distillation again helps structured prediction, but hurts (slightly) the performance on LU tasks from the GLUE benchmark \cite{wang-etal-2018-glue}.

\vspace{1.2mm}
\noindent \textbf{Transformer-Based Dependency Parsing.} Building on the success of preceding neural parsers \cite{chen2014fast,kiperwasser2016simple}, \newcite{dozat2016deep} proposed a biaffine parsing head on top of a Bi-LSTM encoder: contextualized word vectors are fed to two feed-forward networks, producing dependent- and head-specific token representations, respectively. Arc and relation scores are produced via biaffine products between these dependent- and head-specific representation matrices. Finally, the Edmonds algorithm induces the optimal tree from pairwise arc predictions. Most recent DP work \cite{kondratyuk201975,ustun2020udapter} replaces the Bi-LSTM encoder with multilingual BERT's transformer, reporting state-of-the-art parsing performance. \newcite{kondratyuk201975} fine-tune mBERTs parameters on the concatenation of all UD treebanks, whereas \newcite{ustun2020udapter} freeze the original transformer's parameters and inject adapters \cite{houlsby2019parameter} for parsing.  

We propose and work with a simpler transformer-based biaffine parser: we apply biaffine attention directly on representations from transformer's output layer, eliminating the head- and dependendant-based feed-forward mapping. Despite this simplification, our biaffine parser produces DP results comparable to current state-of-the-art parsers.    

\vspace{1.2mm}

\noindent\textbf{Syntactic BERTology.} The substantial body of syntactic probing work shows that BERT \cite{devlin2019bert} (a) encodes text in a hierarchical manner (i.e., it encodes some implicit underlying syntax) \cite{lin2019open}; and (b) captures specific shallow syntactic information (parts-of-speech and syntactic chunks) \cite{tenney2019you,liu2019linguistic}.
\newcite{hewitt2019structural}  find that linear transformations, when applied on BERT's contextualized word vectors, reflect distances in dependency trees. This suggests that BERT encodes sufficient structural information to reconstruct dependency trees (though without arc directionality and relations). \newcite{chi2020finding} extend the analysis to multilingual BERT, finding that its representation subspaces may recover trees also for other languages. They also provide evidence that clusters of head--dependency pairs roughly correspond to UD relations. Similarly, \newcite{kulmizev2020neural} show that BERT's latent syntax corresponds more to UD trees than to shallower SUD \cite{gerdes2018sud} structures. Despite the evident similarity between BERT's latent syntax and formalisms such as UD, there is ample evidence that BERT insufficiently leverages syntax in downstream tasks: it often produces similar predictions for syntactically valid as well as for structurally corrupt sentences (e.g., with random word order) \cite{wallace2019universal,ettinger2020bert,zhao2020limitations}. 

\vspace{1.2mm}

\noindent\textbf{Intermediate Training.} Sometimes called Supplementary Training on Intermediate Labeled-data Tasks (STILT) \cite{phang2018sentence}, intermediate training is a transfer learning setup in which one trains an LM-pretrained transformer on one or more supervised tasks (ideally with large training sets) before final fine-tuning for the target task. \newcite{phang2018sentence} show that intermediate NLI training of BERT on the Multi-NLI dataset \cite{williams2018broad} benefits several language understanding tasks. Subsequent work \cite{wang2019can,pruksachatkun2020intermediate} investigated many combinations of intermediate and target LU tasks, failing to identify any universally beneficial intermediate task. In this work we use DP as an intermediate training task (IPT) for LM-pretrained transformers.

\section{Methodology}

\begin{figure}[t]
    \centering
    \includegraphics[scale=0.5]{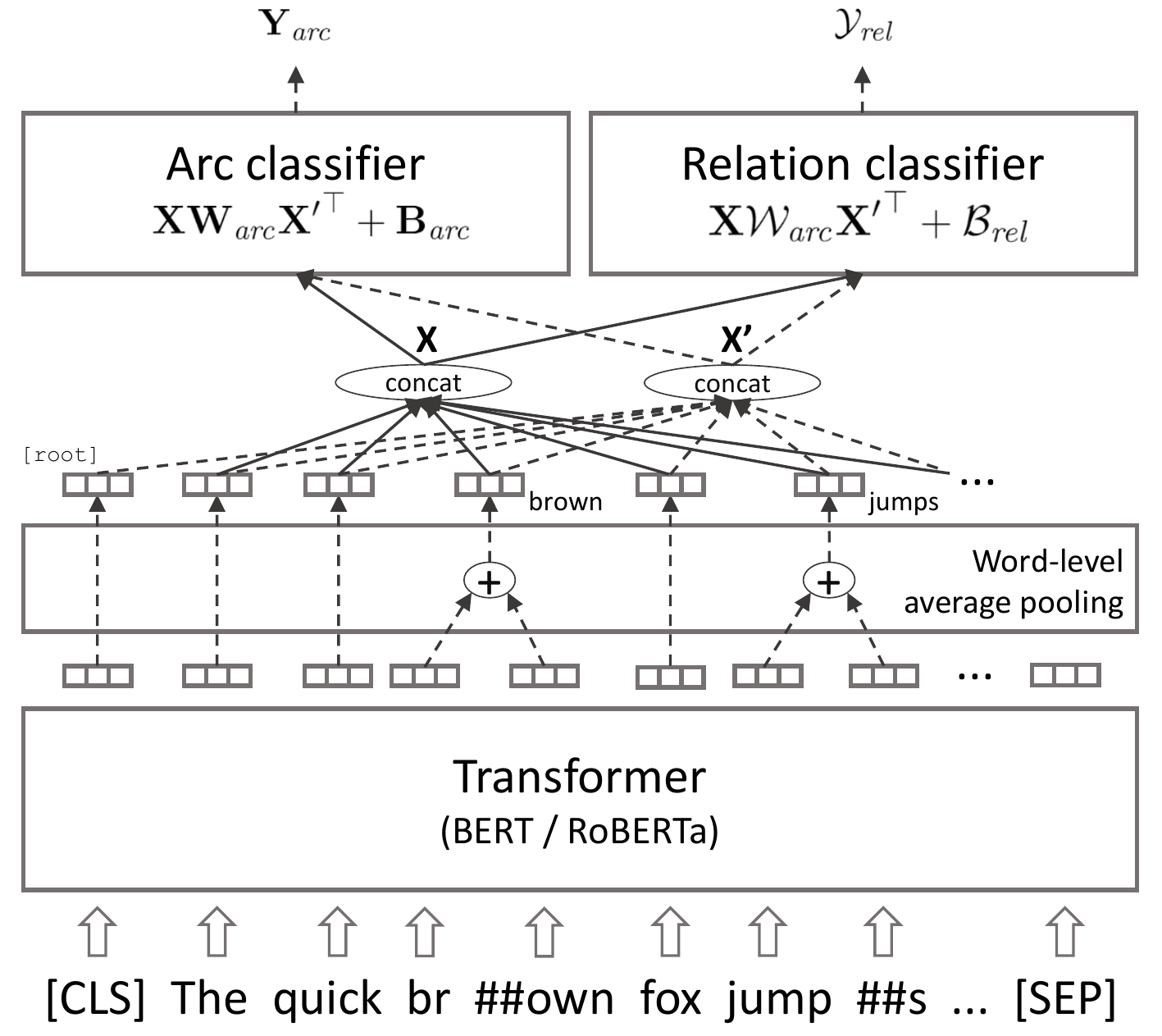}
    \caption{Architecture of our transformer-based biaffine dependency parser.}
    \label{fig:parser}
    \vspace{-1em}
\end{figure}

\noindent \textbf{Biaffine Parser.}
\label{sec:biaffine} 
Our parsing model, illustrated in Figure \ref{fig:parser}, consists of a biaffine attention layer applied directly on the transformer's output (BERT, RoBERTa, mBERT, or XLM-R). We first obtain word-level vectors by averaging transformed representations of their constituent subwords, produced by the transformer. Let $\mathbf{X} \in \mathbb{R}^{N \times H}$ denote the encoding of a sentence with $N$ word-level tokens, consisting of $N$ $H$-dimensional vectors (where $H$ is the transformer's hidden size). We use the transformed representation of the sentence start token (e.g., \texttt{[CLS]} for BERT), $\mathbf{x}_\mathit{CLS} \in \mathbb{R}^H$, as the representation for the \texttt{root} node of the parse tree, and prepend it to $\mathbf{X}$, $\mathbf{X}' = [\mathbf{x}_\mathit{CLS}; \mathbf{X}] \in \mathbb{R}^{(N+1) \times H}$. We then use $\mathbf{X}$ as the representation of syntactic dependants and $\mathbf{X} '$ as the representation of dependency heads. 
We then directly compute the arc and relation scores as biaffine products of $\mathbf{X}$ and $\mathbf{X}'$: 

\vspace{-1em}

{
\small
\begin{equation*}
    \mathbf{Y}_{\mathit{arc}} = \mathbf{X}\mathbf{W}_\mathit{arc}\mathbf{X'}^{\top} + \mathbf{B}_\mathit{arc};\hspace{0.5em}
    \mathbf{Y}_{\mathit{rel}} = \mathbf{X}\mathbf{W}_\mathit{rel}\mathbf{X'}^{\top} + \mathbf{B}_\mathit{rel}
\end{equation*}}

\noindent where $\mathbf{W}_\mathit{arc} \in \mathbb{R}^{H \times H}$ and $\mathbf{W}_\mathit{rel} \in \mathbb{R}^{H \times H \times R}$ denote, respectively, the arc classification matrix and relation classification tensor (with $R$ as the number of relations); $\mathbf{B}_\mathit{arc}$ and $\mathbf{B}_\mathit{rel}$ denote the corresponding bias parameters. We greedily select the dependency head for each word by finding the maximal score in each row of $\mathbf{Y}_{\mathit{arc}}$: while this is not guaranteed to produce a tree, \newcite{zhang2017dependency} show that in most cases it does.\footnote{They also show the performance of greedy decoding to match that of decoding algorithms that produce optimal trees.} Our arc prediction loss is the cross-entropy loss with sentence words (plus the \texttt{root} node) as categorical labels: this implies a different number of labels for different sentences. We compute the relation prediction loss as a cross-entropy loss over gold arcs. Our final loss is the sum of the arc loss and relation loss.

Note that, in comparison with the original biaffine parser \cite{dozat2016deep} and its other transformer-based variants \cite{kondratyuk201975,ustun2020udapter}, we feed word-level representations derived from the transformer's output directly to biaffine products, omitting the dependent- and head-specific MLP transformations. 
Deep task-specific architectures go against the fine-tuning idea: deep transformers have plenty of their own parameters that can be tuned for DP. We want to propagate as much of the explicit syntactic knowledge as possible into the transformer: a deep(er) DP-specific architecture on top of the transformer would impede the propagation of this knowledge to the transformer's parameters.

\vspace{1.2mm}
\noindent \textbf{Downstream Models.}
\label{ss:downstream} 
After IPT, we fine-tune transformers for two \textit{types} of LU tasks: (1) \textit{sequence classification} (SEQC) tasks, where a sequence of text needs to be assigned a discrete label; and (2) \textit{multiple choice classification} (MCC) tasks where we need to select the correct answer between two or more options for a given a premise and/or question. 
For SEQC, we simply apply a softmax classifier on the transformed representation of the sequence start token: $\mathbf{y} = \mathit{softmax}\left(\mathbf{x}_{\mathit{CLS}}\mathbf{W}_\mathit{sc} + \mathbf{b}_\mathit{sc}\right)$ (with $\mathbf{W}_\mathit{sc} \in \mathbb{R}^{H\times C}$ and $\mathbf{b}_\mathit{sc} \in \mathbb{R}^C$ as classifier's parameters and $C$ as the number of task's labels). 

For MCC tasks, we first concatenate each of the offered answer choices (independently of each other) to the premise and/or question, and encode it with the transformer. Since some of these tasks, e.g., COPA \cite{Roemmele:2011aaai,ponti2020xcopa}, have very small training sets, we would like to support model transfer between different MCC tasks. Different multiple-choice classification tasks, however, may differ in the number of choices: a classifier with the number of parameters depending on the number of labels is thus not a good fit; instead, we follow \newcite{Sap:2019siqa} and \newcite{ponti2020xcopa}, and couple the transformer with a feed-forward network outputting a single scalar for each answer. Let $\mathbf{x}^i_\mathit{CLS} \in \mathbb{R}^H$ be the representation of the sequence start token (i.e., \texttt{[CLS]} or \texttt{<s>}) for the concatenation of the premise/question and the $i$-th answer. We obtain the score for the $i$-th answer as follows:

{\small
\begin{equation*}
y_i = \mathbf{W}^o_\mathit{mcc} \tanh \left(\mathbf{W}^h_\mathit{mcc} \mathbf{x}^i_\mathit{CLS} + \mathbf{b}^h_\mathit{mcc}\right)
\end{equation*}}

\vspace{-1em}

\noindent with $\mathbf{W}^h_\mathit{mcc} \in \mathbb{R}^{H \times H}$, $\mathbf{b}^h_\mathit{mcc} \in \mathbb{R}^{H}$ and $\mathbf{W}^o_\mathit{mcc} \in \mathbb{R}^{1 \times H}$ as parameters. We then apply a softmax function on the concatenation of $y_i$ scores of all answers: $\mathbf{y} = \mathrm{softmax}([y_1, \dots, y_K])$, with $K$ as the number of answers (i.e., labels) in the task. Finally, we compute the cross-entropy loss on $\mathbf{y}$.    

\section{Experimental Setup}

We now detail experimental setup, where LU fine-tuning follows Intermediate Parsing Training (IPT).

\subsection{Sequential Fine-Tuning}

Our primary goal is to identify if injection of explicit syntax into transformers via supervised parsing training improves their downstream LU performance -- this translates into sequential fine-tuning: (1) we first attach a biaffine parser from \S\ref{sec:biaffine} on the transformer and train the whole model on a UD treebank; (2) we then couple the syntactically-informed transformer with the corresponding downstream classification head and perform final fine-tuning. We then compare the downstream performance of transformers with and without the IPT step. 

\vspace{1.2mm}

\noindent\textbf{Mono- vs.\,Cross-Lingual IPT Experiments.} In the monolingual setup, we work with English (\en) transformers, BERT and RoBERTa, pretrained on \en corpora. In the zero-shot language transfer setup, where we work with multilingual models, mBERT and XLM-R \cite{Conneau:2020acl}, we first train transformers via IPT on the UD treebank of the target language  (i.e.,\,a language with no downstream training data) before fine-tuning it on the \en training set of the LU task. We experiment with four target languages: German (\de), French (\fr), Turkish (\tr), and Chinese (\zh).\footnote{Selected languages vary in typological and etymological proximity to \en as the source language: \de is in the same (Germanic) branch of Indo-European languages, \fr is from the different branch of the same family, whereas \tr (Turkic) and \zh (Sino-Tibetan) belong to different language families.}

\vspace{1.2mm}

\noindent\textbf{Standard vs.\,Adapter-Based Fine-Tuning.} Standard fine-tuning updates all transformer's parameters, which, for tasks with large training sets may have some drawbacks: (i) fine-tuning may last long and (ii) task-specific information may overwrite the useful distributional knowledge obtained during LM-pretraining. \textit{Adapter-based fine-tuning} \cite{houlsby2019parameter,pfeiffer2020adapterhub} remedies for these potential issues by keeping the original transformer's parameters frozen and inserting new \textit{adapter parameters} in transformer layers. In fine-tuning, both sets of parameters are used to make predictions, but we only update adapters based on loss gradients. As the number of adapter parameters is only a fraction of the number of original parameters (3-8\%), fine-tuning is also much faster. 

Therefore, to account for the possibility of forgetting distributional knowledge in standard IPT fine-tuning, we also carry out adapter-based IPT. 
We follow \newcite{houlsby2019parameter} and inject two \textit{bottleneck adapters} into each transformer layer: first after the multi-head attention sublayer and another after the feed-forward sublayer. In downstream LU tasks, however, we unfreeze the original transformer parameters and fine-tune them together with adapters (now containing syntactic knowledge). 

\subsection{Language Understanding Tasks}

We now outline the downstream LU tasks. For brevity, we report all the technical training and optimization details in the Supplementary Material.

\vspace{1.2mm}

\noindent \textbf{NLI} is a ternary sentence-pair classification task. We predict if the hypothesis is \textit{entailed} by the premise, \textit{contradicts} it, or neither. For monolingual \en experiments, we use Multi-NLI \cite{williams2018broad}. In zero-shot transfer experiments, we train on \en Multi-NLI and evaluate on target language (\de, \fr, \tr, \zh) test portions of the multilingual XNLI dataset \cite{conneau2018xnli}. Models trained on the Multi-NLI datasets have been shown, however, to capture certain heuristics (e.g., lexical overlap) useful for many training instances rather than more complex and generalizable language inference \cite{mccoy2020right}. Because of this, we additionally evaluate  on the HANS dataset \cite{mccoy2020right}, consisting of adversarial examples on which models that capture such heuristics fail.    

\vspace{1.2mm}

\noindent\textbf{Paraphrase Identification} is a binary classification task where we predict if two sentences are mutual paraphrases. For \textsc{en}, we train, validate, and test on respective portions of the PAWS dataset \cite{zhang2019paws}. In zero-shot language transfer, we evaluate on the test \de, \fr, and \zh portions of the PAWS-X dataset \cite{yang2019paws}.  

\vspace{1.2mm}

\noindent\textbf{Commonsense Reasoning.} We evaluate on two multiple-choice classification (MCC) datasets. In monolingual evaluation, we use the SocialIQA (SIQA) dataset \cite{Sap:2019siqa}, testing models' ability to reason about social interactions. Each SIQA instance consists of a premise, a question, and three possible answers. For zero-shot language transfer experiments, we resort to the recently published XCOPA dataset \cite{ponti2020xcopa}, obtained by translating test portions of the \en COPA (Choice of Plausible Alternatives) dataset \cite{Roemmele:2011aaai} to 11 languages. As mentioned, (X)COPA is an MCC task, with each instance containing a premise, a question,\footnote{While SIQA has unconstrained questions, (X)COPA has only two question types: a) What is the CAUSE of this (premise)? and b) What is the RESULT of this (premise)?} and two possible answers. Due to the very limited size of the \en COPA training set (mere 400 instances), we follow \newcite{ponti2020xcopa} and evaluate the models fine-tuned on SIQA (\en) on the XCOPA test portions (in \tr and \textsc{zh}).      

\setlength{\tabcolsep}{6.5pt}
\def\arraystretch{0.83}
\begin{table*}[t!]
\centering
\small
\begin{tabularx}{\linewidth}{l l cc cc cc cc cc} 
\toprule
 & & \multicolumn{2}{c}{\en (EWT)} & \multicolumn{2}{c}{\de (GSD)} & \multicolumn{2}{c}{\fr (GSD)} & \multicolumn{2}{c}{\tr (IMST)} & \multicolumn{2}{c}{\zh (GSD)} \\ \cmidrule(lr){3-4}\cmidrule(lr){5-6}\cmidrule(lr){7-8}\cmidrule(lr){9-10}\cmidrule(lr){11-12}
\textbf{Transformer} & \textbf{Fine-tune} & UAS & LAS & UAS & LAS & UAS & LAS & UAS & LAS & UAS & LAS \\ \midrule
\multirow{2}{*}{BERT} & Standard & 91.9 & 89.3 & -- & -- & -- & -- & -- & -- & -- & -- \\
& Adapter & 90.1 & 87.3 & -- & -- & -- & -- & -- & -- & -- & -- \\ \cdashline{2-12}
\multirow{2}{*}{RoBERTa} & Standard & 93.0 & \textbf{90.5} & -- & -- & -- & -- & -- & -- & -- & --  \\
& Adapter & 91.5 & 88.7 & -- & -- & -- & -- & -- & -- & -- & -- \\ \midrule 
\multirow{2}{*}{mBERT} & Standard & 91.5 & 88.9 & 76.3 & 72.0 & 94.1 & 91.3 & 75.5 & 67.5 & 87.0 & \textbf{83.8} \\
& Adapter & 89.6 & 86.8 & 75.1 & 70.1 & 92.8 & 89.7 & 66.4 & 57.8 & 81.0 & 77.4 \\ \cdashline{2-12} 
\multirow{2}{*}{XLM-R} & Standard & \textbf{93.1} & \textbf{90.5} & \textbf{89.4} & \textbf{85.0} & \textbf{94.3} & \textbf{91.7} & \textbf{77.9} & \textbf{70.0} & 79.0 & 75.6 \\
& Adapter & 91.4 & 88.6 & 88.3 & 83.8 & 93.1 & 90.3 & 72.1 & 64.1 & 73.8 & 70.3 \\ \midrule
\multicolumn{2}{c}{\textbf{Baseline}: UDify~(mBERT, Standard)} & 91.0 & 88.5 & 87.8 & 83.6 & 93.6 & 91.5 & 74.6 & 67.4 & \textbf{87.9} & \textbf{83.8} \\
\bottomrule
\end{tabularx}
\caption{Dependency parsing performance of our transformer-based biaffine parsers. 
}
\vspace{-1.5mm}
\label{tbl:parsing}
\end{table*}

\subsection{Training and Optimization Details} 
\label{ssec:optimization}

All the transformer models with which we experiment -- \en BERT, mBERT, \en RoBERTa, and XLM-R have $L = 12$ layers and hidden representations of size $H = 768$. We apply a dropout ($p = 0.1$) on the transformer outputs before forwarding them to the task-specific classification heads (i.e., biaffine parsing head in intermediate parsing training, and MCC or SEQC heads in downstream fine-tuning). We optimize the parameters using the Adam algorithm \cite{kingma2015adam}: we found the initial learning rate of $10^{-5}$ to offer stable convergence in both intermediate parsing training and downstream fine-tuning for all LU tasks. We train for at most $30$ epochs over the respective training set, with early stopping based on the development loss.\footnote{We measure the development loss every $U$ update steps and stop the training if the loss does not decrease over $10$ consecutive measurements. We set $U = 500$ in NLI training and $U = 250$ in all other training procedures.} On UD treebanks and SIQA we train in batches of size $8$, whereas on Multi-NLI and PAWS we train in batches of size $32$. In Adapter-based IPT, we set the adapter size to $64$ and use GELU \cite{hendrycks2016gaussian} as the activation function in adapter layers.

\setlength{\tabcolsep}{4.8pt}
\begin{table}[t!]
\centering
\def\arraystretch{0.83}
\small
\begin{tabularx}{\linewidth}{ll cccc} 
\toprule
\textbf{Transf.} & \textbf{Parsing FT} & NLI & HANS & PAWS & SIQA \\ \midrule
\multirow{3}{*}{BERT} & None & 84.1 & 53.3 & \textbf{92.4} & \textbf{60.7} \\ \cdashline{2-5}
& Standard & \textbf{84.4} & \textbf{56.7} & 91.9 & 58.8 \\
& Adapter & 84.1 & 53.3 & \textbf{92.4} & 58.3 \\ \midrule
\multirow{3}{*}{RoBERTa} & None & \textbf{88.4} & \textbf{67.4} & 94.7 & 67.2\\ \cdashline{2-5}
& Standard & 87.7 & 64.5 & \textbf{94.9} & 66.5 \\
& Adapter & 87.9 & 66.3 & 94.7 & \textbf{67.3} \\
\bottomrule
\end{tabularx}
\caption{Downstream LU performance of monolingual \en transformers (BERT and RoBERTa). \textbf{None}: no IPT; \textbf{Standard}: IPT via standard fine-tuning; \textbf{Adapter}: IPT via adapter-based fine-tuning.}
\label{tbl:mono}
\vspace{-1em}
\end{table}
\setlength{\tabcolsep}{10pt}
\begin{table*}[t!]
\centering
\def\arraystretch{0.83}
\small
\begin{tabularx}{\linewidth}{l l cccc ccc cc} 
\toprule
 & & \multicolumn{4}{c}{XNLI} & \multicolumn{3}{c}{PAWS-X} & \multicolumn{2}{c}{XCOPA} \\ \cmidrule(lr){3-6}\cmidrule(lr){7-9}\cmidrule(lr){10-11}
\textbf{Transformer} & \textbf{Parse FT} & \de & \fr & \tr & \zh & \de & \fr & \zh & \tr & \zh \\ \midrule
\multirow{3}{*}{mBERT} & None & 71.0 & 73.7 & \textbf{63.0} & 70.3 & 85.1 & 86.3 & 76.4 & 52.0 & 61.2 \\ \cdashline{2-11}
& Standard & 71.4 & 72.9 & 61.5 & \textbf{70.4} & 85.4 & 86.9 & \textbf{79.8} & \textbf{57.4} & \textbf{65.4} \\
& Adapter & \textbf{71.7} & \textbf{74.8} & 62.5 & 70.2 & \textbf{85.8} & \textbf{87.1} & 78.7 & 50.4 & 61.6 \\ \midrule
\multirow{3}{*}{XLM-R} & None & 77.1 & \textbf{78.1} & 73.4 & 73.8 & \textbf{88.3} & \textbf{89.3} & \textbf{81.4} & \textbf{61.2} & 66.4 \\ \cdashline{2-11}
& Standard & 76.1 & 77.2 & 73.1 & 73.8 & 86.4 & 89.2 & 81.1 & 59.2 & \textbf{67.4} \\
& Adapter & \textbf{77.8} & 76.4 & \textbf{73.9} & \textbf{74.7} & 86.7 & 88.7 & 80.7 & 57.4 & 65.6 \\
\bottomrule
\end{tabularx}
\caption{Performance of multilingual transformers, mBERT and XLM-R, in zero-shot language transfer for downstream LU tasks, with and without prior intermediate dependency parsing training on target language treebanks.}
\vspace{-1.5mm}
\label{tbl:xling}
\end{table*}

\section{Evaluation}

We first discuss parsing performance of our novel biaffine parser (see \S\ref{sec:biaffine}). We then show transformers' downstream LU performance after IPT, both in monolingual \en setting and in zero-shot  transfer.

\subsection{Results and Discussion} 

\noindent \textbf{Parsing Performance.} In order to judge the benefits of IPT in downstream LU, we must first verify parsing performance of our biaffine parser, i.e., that we successfully fine-tune transformers for DP. 
Table \ref{tbl:parsing} shows that our biaffine parser gives state-of-the-art performance for all five languages in our study. Our (m)BERT-based parser outperforms UDify \cite{kondratyuk201975}, also based on mBERT, for \en, \fr, and \tr, and performs comparably for \zh.\footnote{Our mBERT-based parser performs poorly for \de: the cause of it is unclear and this requires further investigation.} Our parser based on XLM-R additionally yields an improvement over UDify for \de as well. 
It is worth noting that UDify trains the mBERT-based parser (1) on the concatenation of all UD treebanks and that it (2) additionally exploits gold UPOS and lemma annotations. We train our parsers only on the training portion of the respective treebank without using any additional morpho-syntactic information.\footnote{Also, since absolute parsing performance is not the primary objective of this work, we did not perform extensive language-specific hyperparameter tuning. One could likely obtain better parsing scores than what we report in Table~\ref{tbl:parsing} with careful language-specific model selection.}
Our mBERT-based parser outperforms our XLM-R-based parser only for \zh: this is likely due to a tokenization mismatch between XLM-R's subword tokenization for \zh and gold tokenization in the \zh-GSD treebank.\footnote{We explain this mismatch in the Appendix.}

\vspace{1.2mm}

\noindent\textbf{Monolingual \en Results.} Table \ref{tbl:mono} quantifies the effects of applying IPT with the \en-EWT UD treebank to BERT and RoBERTa. We report downstream LU performance on NLI, PAWS, and SIQA. The reported results do not favor supervised parsing (i.e., explicit syntax): compared to original transformers that have not been exposed to any explicit syntactic supervision, variants exposed to UD syntax via IPT (Standard, Adapter) fail to produce any significant gains for any of the downstream LU tasks. One cannot argue that the cause of this might be forgetting (i.e., overwriting) of the distributional knowledge obtained in LM pretraining during IPT: Adapter IPT variants, in which all distributional knowledge is preserved by design, also fail to yield any significant LU gains. IPT yields the largest gain (+3.4\%) for BERT on HANS -- the NLI dataset consisting of adversarial examples for which syntax deliberately affects the sentence meaning more directly. The same effect, however, is not there for RoBERTa, suggesting that the additional syntactic knowledge that BERT gets through IPT, RoBERTa seems to obtain through larger-scale pretraining.   

\vspace{1.2mm}

\noindent\textbf{Zero-Shot Language Transfer.} We show the results obtained for zero-shot downstream language transfer setup, for both mBERT and XLM-R, in Table~\ref{tbl:xling}. Again, these results do not particularly favor the intermediate injection of explicit syntactic information in general. However, in few cases we do observe gains from the intermediate target-language parsing training: e.g., 3\% gain on PAWS-X for \zh as well as 4\% and 5\% gains on XCOPA for \zh and \tr, respectively. Interestingly, all substantial improvements are obtained for mBERT; for XLM-R, the improvements are less consistent and less pronounced. This might be due to XLM-R's larger capacity which makes it less susceptible to the ``curse of multilinguality'' \cite{Conneau:2020acl}: with the subword vocabulary twice as large as mBERT's, XLM-R is able to store more language-specific information. Also, XLM-R has seen substantially more target language data in LM-pretraining than mBERT for each language. This might mean that the larger IPT gains for mBERT come from mere exposure to additional target language text rather than from injection of explicit syntactic UD signal (see further analyses in \S\ref{sec:further}).

\subsection{Further Analysis and Discussion}
\label{sec:further}

We first compare the impact of IPT with the effect of additional LM training on the same raw data. We then quantify the topological modification that IPT makes in transformers' representation spaces.

\vspace{1.2mm}

\noindent\textbf{Explicit Syntax or Just More Language Data?} We scrutinize the IPT gains that we observe in some zero-shot language transfer experiments. We hypothesize that these gains may, at least in part, be credited to transformer simply seeing more target language data. 
To investigate this, we replace IPT with intermediate (masked) language modeling training (ILMT) on the same data (i.e., sentences from the respective treebank used in IPT) before final downstream LU fine-tuning.
Because MLM is a self-supervised objective, we can credit all differences in downstream LU performance between ILMT and IPT variants of the same pretrained transformer to supervised parsing, i.e., to the injection of explicit UD knowledge.

\begin{figure}
    \begin{center}
    \includegraphics[scale=0.40]{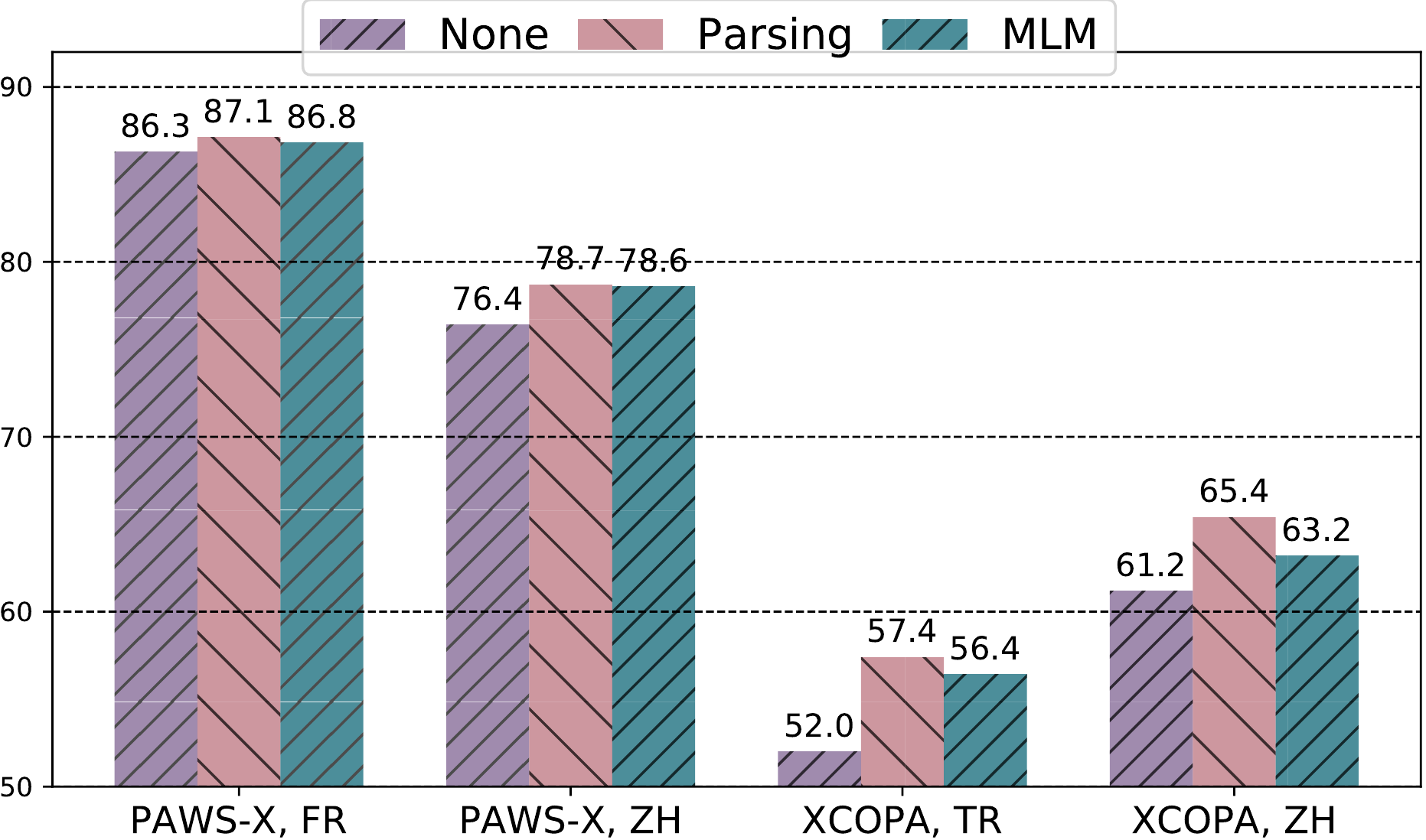}
    \caption{Comparison of IPT and ILMT in zero-shot language transfer experiments with mBERT on PAWS-X (\fr and \zh) and XCOPA (\tr and \zh). \textbf{None}: no intermediate training; \textbf{Parsing}: intermediate parsing training; \textbf{MLM}: intermediate masked LM training.}
    \label{fig:pabl}
    \end{center}
    \vspace{-1.5em}
\end{figure}
\begin{figure*}[t!]
    \centering
    \includegraphics[scale=0.58]{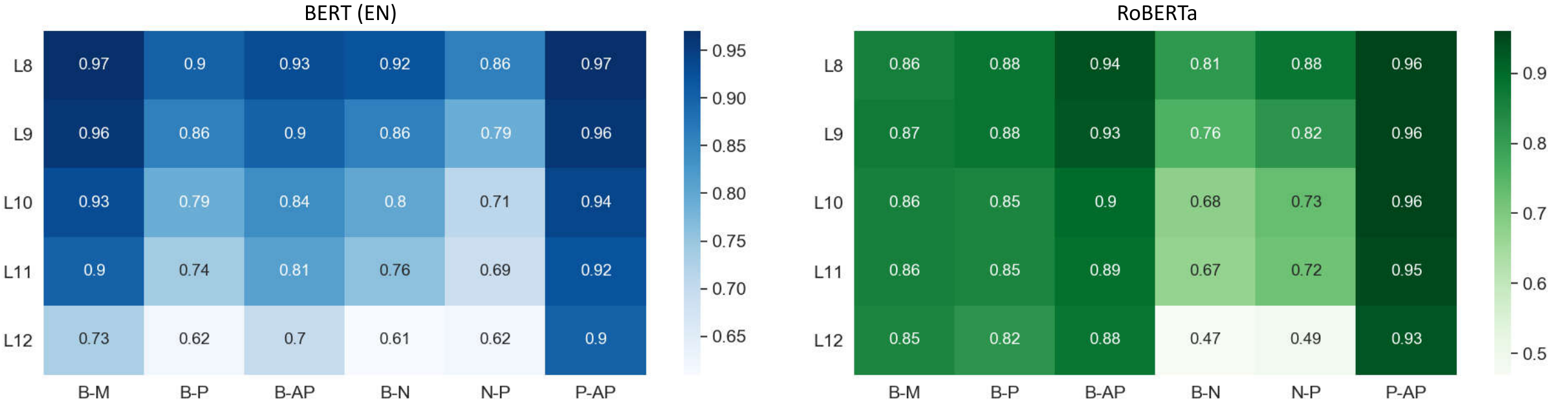}
    \caption{Topological similarity (l-CKA) for pairs of BERT and RoBERTa variants, before and after different fine-tuning steps (B, M, P, AP, and N). \textbf{Rows}:\,transformer layers;\,\textbf{Columns}:\,pairs of transformer variants in comparison.}
    \label{fig:hm_en}
    \vspace{-1em}
\end{figure*}

\vspace{1.2mm}
\noindent\textbf{ILMT Details.} We mask 15\% of subword tokens in each sentence and predict them with a linear classifier applied on transformed representations of \texttt{[MASK]} tokens. We compute the cross-entropy loss and use the same hyperparameter configuration as described in \S\ref{ssec:optimization}. The development set, used for early stopping, is subdued to fixed masking, whereas we mask the training sentences dynamically, before feeding them to the transformer.   

\vspace{1.2mm}
\noindent\textbf{Results.} We run this analysis for setups in which we observe substantial gains from IPT: PAWS-X for mBERT (\textit{Adapter} fine-tuning, for \fr and \zh) and XCOPA for mBERT (\textit{Standard} fine-tuning, \tr and \zh). The comparison between IPT and ILMT for these setups is provided in Figure~\ref{fig:pabl}. 
Like IPT, ILMT on mBERT generates downstream gains over direct downstream fine-tuning (i.e., no intermediate training) in all four setups. The gains from ILMT (with the exception of XCOPA for \zh) are almost as large as gains from IPT. This suggests that most of the gain with IPT comes from seeing more target language text, and prevents us from concluding that the explicit syntactic annotation is responsible for the LU improvements in zero-shot downstream transfer. This interpretation is corroborated by the fact that IPT gains roughly correlate with the amount of language-specific data seen in LM-pretraining: the gains are more prominent for mBERT than for XLM-R and for \tr and \zh than for \fr and \de (see Table~\ref{tbl:xling}).

\vspace{1.2mm}

\begin{figure*}[t!]
    \centering
    \includegraphics[scale=0.78]{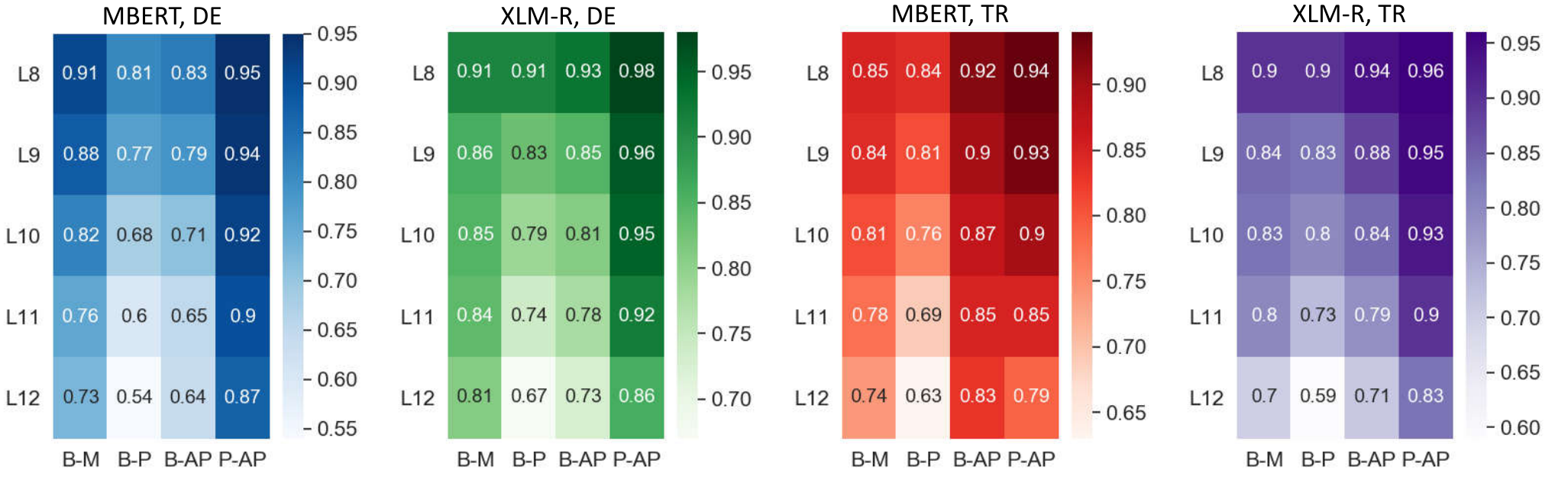}
    \caption{Analysis of topological similarity (l-CKA) for variants of mBERT and XLM-R before and after IPT and ILMT (B, M, P, AP) in zero-shot transfer experiments. Results shown for intermediate parsing on \de and \tr data.}
    \label{fig:hm_xling}
    \vspace{-1em}
\end{figure*}

\noindent\textbf{Changes in Representation Spaces.} Finally, we analyze how fine-tuning transformers on different tasks modifies the topology of their representation spaces. We encode the set of sentences $S$ from the test portions of treebanks used in IPT\footnote{IPT itself only consumes train and development portions of UD treebanks. We can thus safely use sentences from test portions in this analysis, without risking information leakage.} with different variants: (a) Base (B): original LM-pretrained transformer, no further training; (b) MLM (M): after ILMT; (c) Parsing (P): after Standard IPT; and (d) Adapter-Parsing (AP): after Adapter-based IPT; for monolingual transformers (BERT and RoBERTa), also with (e) NLI (N): after NLI fine-tuning (without any intermediate training). 
We analyze the representations in each transformer layer separately: we represent each sentence $s \in S$ with the average of subword vectors from that layer (excluding sequence start and end tokens). Let $\mathbf{X}_1\text{ and }\mathbf{X}_2 \in \mathbb{R}^{|S| \times H}$ contain corresponding representations of sentences from $S$ from the $i$-th layer of two transformer variants (e.g., B and P). We measure the topological similarity of the $i$-th layers of the two transformers with the linear centered kernel alignment (l-CKA) \cite{kornblith2019similarity}:\footnote{$\mathbf{X}_1$ and $\mathbf{X}_2$ must first be column-wise mean-centered.} 

\vspace{-0.5em}

{\footnotesize
\begin{equation*}
l\text{-}\mathit{CKA}(\mathbf{X}_1, \mathbf{X}_2) = \frac{\norm{\mathbf{X}^{\top}_2\mathbf{X}_1}^2_F}{\left(\norm{\mathbf{X}^{\top}_1\mathbf{X}_1}_F\right) \left(\norm{\mathbf{X}^{\top}_2\mathbf{X}_2}_F\right)}.
\end{equation*}
}

\vspace{-0.5em}


\noindent Although not invariant to all linear transformations, l-CKA is invariant to orthogonal projection and isotropic scaling, which suffices for our purposes. 
We base our analysis on the following assumption: the extent of change in transformers' representation space topology (reflected by l-CKA), is proportional to the novelty of knowledge injected in fine-tuning. Put differently, injection of new (i.e., missing) knowledge should substantially change the topology of the space (low l-CKA score).    

Figure \ref{fig:hm_en} shows the heatmap of l-CKA scores for pairs of BERT and RoBERTa variants, for layers L8-L12.\footnote{Most l-CKA scores in layers L1-L7 are very high ($\ge 0.9$) and provide little insight. See the Supplementary Material.} 
Comparing B-P and B-N reveals that IPT changes the topology of BERT's higher layers roughly as much as NLI fine-tuning does, implying that both the English UD treebank (\en-EWT) and Multi-NLI data contain a non-negligible amount of novel knowledge for BERT. However, the direct N-P comparison shows that IPT and NLI enrich BERT (also RoBERTa) with different type of knowledge, i.e., they change the representation spaces of its layers in different ways. This suggests that the transformers cannot acquire the missing knowledge needed for NLI from IPT (i.e., from \en-EWT), and explains why IPT is not effective for NLI. 

IPT (comparison B-P) injects more new information than ILMT (comparison B-M), and this is more pronounced for BERT than for RoBERTa. IPT and ILMT change RoBERTa's parameters much less than BERT's (see B-M and B-P l-CKA scores for L11/L12), which we interpret as additional evidence, besides RoBERTa consistently outscoring BERT, that RoBERTa encodes richer language representations, due to its larger-scale and longer training. It also agrees with suggestions that BERT is ``undertrained'' for its capacity \cite{liu_roberta_2019}. 

Very high B-P (and B-AP) l-CKA scores in lower layers suggest that the explicit syntactic knowledge from human-curated treebanks is redundant w.r.t. the structural language knowledge transformers obtain through LM pretraining. This is consistent with concurrent observations \cite{chi2020finding,kulmizev2020neural} showing (some) correspondence between structural knowledge of (m)BERT and UD syntax.     
Finally, we observe highest l-CKA scores in the P-AP column, suggesting that Standard and Adapter IPT inject roughly the same syntactic information, despite different fine-tuning mechanisms. 




Figure \ref{fig:hm_xling} illustrates the results of the same analysis for language transfer experiments, for \de and \tr (scores for \fr and \zh are in the Appendix). The effects of ILMT and IPT (B-M, B-P/B-AP) for \de and \tr with mBERT and XLM-R resemble those for \en with BERT and RoBERTa: ILMT changes transformers less than IPT. The amount of new syntactic knowledge  IPT injects is larger (l-CKA scores are lower) than for \en, especially for XLM-R (vs.\,RoBERTa for\,\en). We believe that it reflects the relative under-representation of the target language in the model's multilingual pretraining corpus (e.g., for \textsc{tr}): this leads to poorer representations of target language structure by mBERT and XLM-R compared to BERT's and RoBERTa's representation of \en structure. This gives us two seemingly conflicting empirical findings: (a) IPT appears to inject a fair amount of target-language UD syntax, but (b) this translates to (mostly) insignificant and inconsistent gains in language transfer in LU tasks (especially so for XLM-R, cf.\,Table \ref{tbl:xling}). A plausible reconciling hypothesis is that there is a substantial mismatch between the type of structural information we obtain through supervised (UD) parsing and the type of structural knowledge beneficial for LU tasks. If true, this hypothesis would render supervised parsing rather unavailing for high-level language understanding, at least in the context of LM-pretrained transformers, the current state of the art in NLP. This warrants further investigation, and we hope that our work will inspire further discussion and additional studies. 


%
\section{Conclusion}

We thoroughly examined the effects of leveraging formalized syntactic structures (UD) in state-of-the-art neural language models (e.g., RoBERTa, XLM-R) for downstream language understanding (LU) tasks, both in monolingual and language transfer settings. 
The key results, obtained through intermediate parsing training (IPT) based on a state-of-the-art-level dependency parser, indicate that explicit syntax, at least in our extensive experiments, provides negligible impact on LU tasks. 

Besides offering extensive empirical evidence of the mismatch between explicit syntax and improved LU performance with state-of-the-art transformers, this study sheds new light on some fundamental questions such as the one in the title. Similar to word embeddings \cite{mikolov2013distributed} removing sparse lexical features from the NLP horizon, will transformers make supervised parsing obsolete for LU applications or not? More dramatically, in the words of Rens \newcite{bod2007end}: ``Is the end of supervised parsing in sight'' for semantic LU tasks?\footnote{The answer is 'Probably no': formalized syntactic structures will still be an important source of inductive bias, especially in setups without sufficient text data for large-scale pretraining; our experiments, however, validate that state-of-the-art transformer models can implicitly capture that inductive bias in high-resource setups and for major languages.}





\section*{Acknowledgments}
We thank the anonymous reviewers (especially R2!) for the exceptionally meaningful and helpful comments. Goran Glava\v{s} is supported by the Baden W\"{u}rttemberg Stiftung (Eliteprogramm, AGREE grant). The work of Ivan Vuli\'{c} is supported by the ERC Consolidator Grant LEXICAL: Lexical Acquisition Across Languages (no. 648909).

\bibliographystyle{acl_natbib}
\bibliography{references}

\clearpage

\appendix
\setlength{\tabcolsep}{5pt}
\begin{table*}
\def\arraystretch{0.93}
\centering
{\footnotesize
\begin{tabularx}{\textwidth}{l l l l X}
\toprule
{\bf Name} & {\bf Lang} & \textbf{Vocab} & {\bf Params} & {\bf URL} \\ \midrule
BERT & {\en} & 29K & 110M & {\url{https://huggingface.co/bert-base-cased}} \\
RoBERTa & {\en} & 50K & 110M & {\url{https://huggingface.co/roberta-base}} \\
mBERT & {Multiling.} & 119K & 125M & {\url{https://huggingface.co/bert-base-multilingual-cased}} \\
XLM-R & {Multiling.} & 250K & 125M & {\url{https://huggingface.co/xlm-roberta-base}} \\
\bottomrule
\end{tabularx}
}
\vspace{-1.5mm}
\caption{LM-pretrained transformer models used in our study.}
\label{tbl:models}
\end{table*}
\setlength{\tabcolsep}{8pt}
\begin{table*}
\centering
{\footnotesize
\begin{tabularx}{0.6\textwidth}{ll XXX}
\toprule
{\bf Lang} & \textbf{Treebank} & \textbf{Train} & \textbf{Dev} & \textbf{Test} \\ \midrule
{\en} & EWT & 12,538 & 2,002 & 2,077 \\
{\de} & GSD & 13,810 & 799 & 977 \\
{\fr} & GSD & 14,440 & 1,475 & 416 \\ 
{\tr} & IMST & 3,664 & 988 & 983 \\
{\zh} & GSD & 3,996 & 500 & 500 \\
\bottomrule
\end{tabularx}
}
\vspace{-1.5mm}
\caption{Universal Dependencies treebanks used in our study. We display sizes of train, development, and test portions in terms of number of sentences.}
\label{tbl:treebanks}
\end{table*}
\setlength{\tabcolsep}{6pt}
\begin{table*}
\def\arraystretch{0.93}
\centering
{\footnotesize
\begin{tabularx}{\textwidth}{l l X}
\toprule
{\bf Task} & {\bf Dataset} & {\bf URL} \\ \midrule
Natural Language Inference & Multi-NLI & \url{https://cims.nyu.edu/~sbowman/multinli} \\
Natural Language Inference & XNLI & \url{https://github.com/facebookresearch/XNLI} \\
Paraphrase identification & PAWS(-X) & \url{https://github.com/google-research-datasets/paws} \\
Commonsense social reasoning & SIQA & \url{https://maartensap.github.io/social-iqa} \\
Commonsense causal reasoning & COPA & \url{https://people.ict.usc.edu/~gordon/copa.html} \\
Commonsense causal reasoning & XCOPA & \url{https://github.com/cambridgeltl/xcopa} \\
\bottomrule
\end{tabularx}
}
\vspace{-1.5mm}
\caption{Links to downstream language understanding datasets used in our work. }
\label{tbl:datasets}
\end{table*}

\section{Reproducibility}

We first provide details on where to obtain datasets and code used in this work.   

\subsection{Datasets}

Table \ref{tbl:treebanks} lists the sizes (in number of sentences) of Universal Dependencies treebanks that we use for our intermediate parsing training and evaluation of our biaffine dependency parsers. The UD treebanks v.2.5, which we used in this work, are available at: \url{http://hdl.handle.net/11234/1-3105}. 
In Table \ref{tbl:datasets} we provide links to language understanding datasets used in our study. 

\subsection{Code and Dependencies}

We make our code available at: \url{https://github.com/codogogo/parse_stilt}. Our code is built on top of the HuggingFace Transformers framework: \url{https://github.com/huggingface/transformers} (v.\,2.7). Table \ref{tbl:models} details the LM-pretrained transformer models from this framework which we exploited in this work. Besides the Transformers library, our code only relies on standard Python's scientific computing libraries (e.g., \texttt{numpy}).

\section{\zh Tokenization: XLM-R vs.\,GSD}
\begin{CJK*}{UTF8}{bsmi}
A word-level token from the parse tree normally corresponds to one or more transformer's subword tokens: we thus average subword vectors to obtain word vectors for biaffine parsing. For XLM-R and the \zh GSD treebank, however, a single XLM-R's subword token often corresponds to two treebank tokens. E.g., the sequence ``只是二選一做決擇'' with treebank tokenization [`只', `是', `二', `選', `一', `做', `決擇'] is tokenized as [`只是', `二', `選', `一', `做', `決', `擇'] by XLM-R. Two treebank tokens, `只' and `是', are captured with a single XLM-R ``subword'' token, `只是'. To ensure that each XLM-R subword token corresponds to exactly one treebank token, we inject spaces between treebank tokens before XLM-R tokenization: we then obtain the subword tokenization [`只', `是', `二', `選', `一', `做', `決, `擇']. However, this is suboptimal for XLM-R:\,its representations of tokens '只' and '是' are probably less reliable than that of the '只是' token. We believe this is why mBERT (without tokenization mismatches for \zh) outperforms XLM-R in \zh parsing.     

\end{CJK*}

\section{Complete Topology Analysis Results}

\begin{figure*}
    \centering
    \includegraphics[scale=0.65]{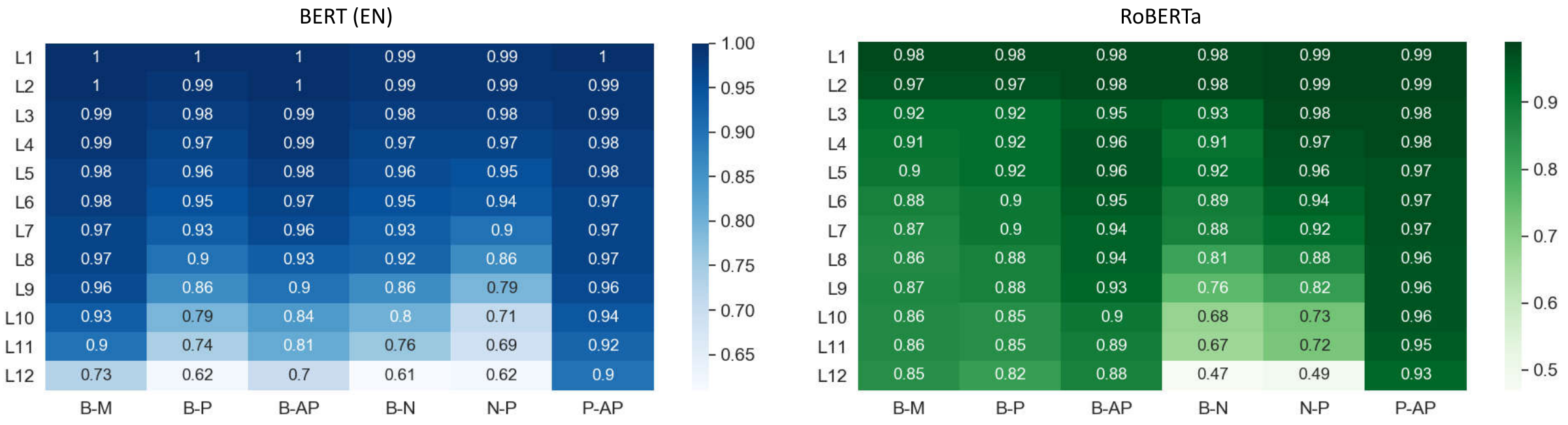}
    \caption{Full results of the topological similarity analysis (l-CKA) for pairs of BERT and RoBERTa variants, before and after different fine-tuning steps (B, M, P, AP, and N). \textbf{Rows}:\,transformer layers;\,\textbf{Columns}:\,pairs of transformer variants in comparison.}
    \label{fig:full_en}
\end{figure*}

\begin{figure*}
    \centering
    \includegraphics[scale=0.72]{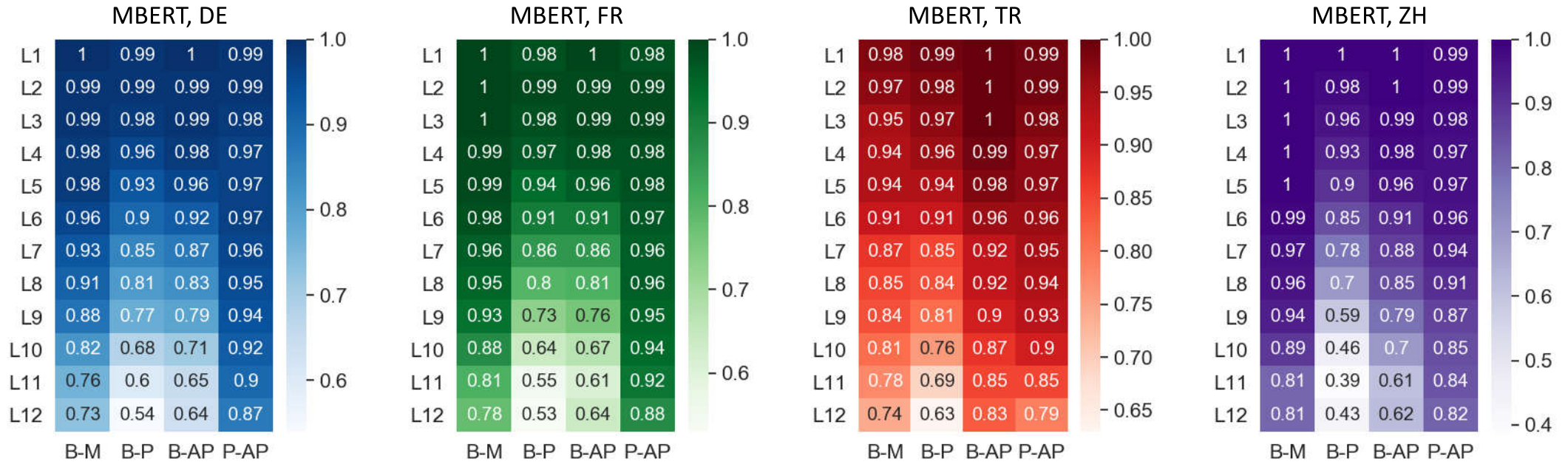}
    \caption{Full results of the topological similarity analysis (l-CKA) for variants of mBERT before and after IPT and ILMT (B, M, P, AP) for the following target languages (left to right): \de, \fr, \tr, and \zh.}
    \label{fig:mbert_full}
\end{figure*}

\begin{figure*}[ht]
    \centering
    \includegraphics[scale=0.72]{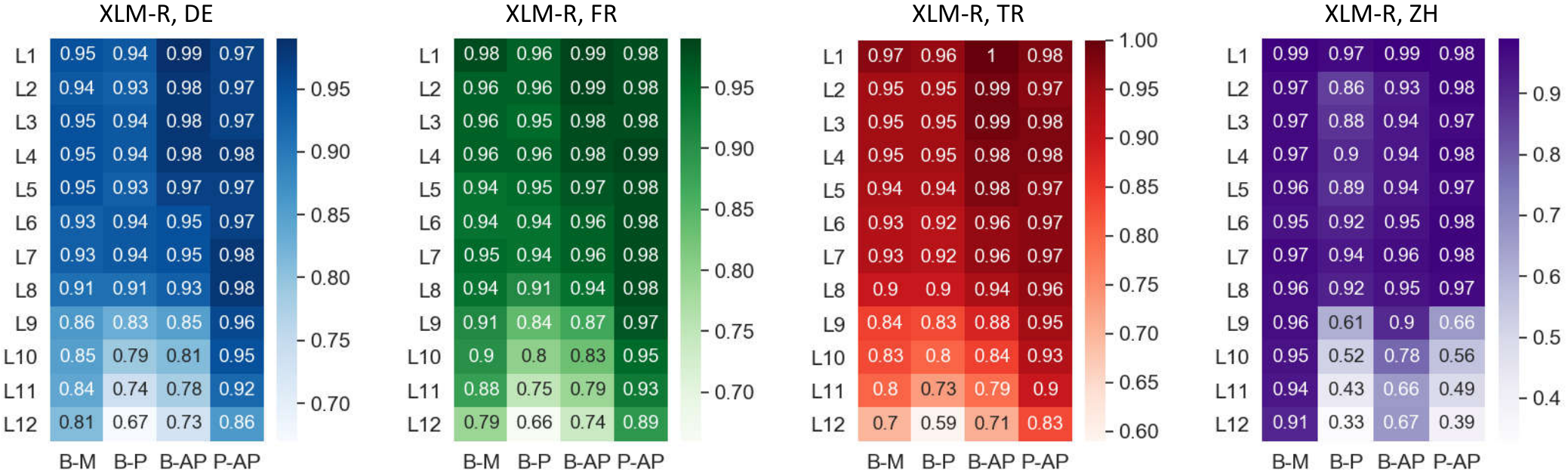}
    \caption{Full results of the topological similarity analysis (l-CKA) for variants of XLM-R before and after IPT and ILMT (B, M, P, AP) for the following target languages (left to right): \de, \fr, \tr, and \zh.}
    \label{fig:xlmr_full}
\end{figure*}

Finally, we show the complete results (for all layers, all transformers, and all languages covered in our experiments) of our topological analysis of transformers' representations before and after different fine-tuning steps. Figure \ref{fig:full_en} shows the analysis results for monolingual \en transformers, BERT and RoBERTa. Figure \ref{fig:mbert_full} and Figure \ref{fig:xlmr_full} show the results for multilingual transformers, mBERT and XLM-R, respectively, for all four target languages included in our experiments: \de, \fr, \tr, and \zh. 

\end{document}